\documentclass[runningheads]{llncs}
\usepackage{graphicx}
\usepackage{gensymb}
\usepackage{multirow}
\usepackage{cite}
\usepackage{amsmath}
\usepackage{amssymb}
\usepackage{hyperref}
\hypersetup{
    colorlinks = true,
    linkcolor= magenta,
}
\newcommand{\R}{\mathbb{R}}

\begin{document}

\title{RadarFormer: Lightweight and Accurate Real-Time Radar Object Detection Model}
\titlerunning{RadarFormer}
\author{Yahia Dalbah \inst{1}\orcidID{0000-0003-1488-4794} \and
Jean Lahoud\inst{1}\orcidID{0000-0003-0315-6484} \and
Hisham Cholakkal\inst{1}\orcidID{0000-0002-8230-9065}}

\authorrunning{Y. Dalbah et al.}

\institute{Mohamed Bin Zayed University of Artificial Intelligence, UAE 
\email{\{yahia.dalbah,jean.lahoud,hisham.cholakkal\}@mbzuai.ac.ae}}

\maketitle             
\begin{abstract}

The performance of perception systems developed for autonomous driving vehicles has seen significant improvements over the last few years. This improvement was associated with the increasing use of LiDAR sensors and point cloud data to facilitate the task of object detection and recognition in autonomous driving. However, LiDAR and camera systems show deteriorating performances when used in unfavorable conditions like dusty and rainy weather. Radars on the other hand operate on relatively longer wavelengths which allows for much more robust measurements in these conditions. Despite that, radar-centric data sets do not get a lot of attention in the development of deep learning techniques for radar perception. In this work, we consider the radar object detection problem, in which the radar frequency data is the only input into the detection framework. We further investigate the challenges of using radar-only data in deep learning models. We propose a transformers-based model, named RadarFormer, that utilizes state-of-the-art developments in vision deep learning. Our model also introduces a channel-chirp-time merging module that reduces the size and complexity of our models by more than 10 times without compromising accuracy. 
Comprehensive experiments on the CRUW radar dataset demonstrate the advantages of the proposed method. Our RadarFormer performs favorably against the state-of-the-art methods while being 2x faster during inference and requiring only one-tenth of their model parameters. The code associated with this paper is available at \href{https://github.com/YahiDar/RadarFormer}{https://github.com/YahiDar/RadarFormer}.

\end{abstract}

\hypersetup{
    colorlinks = False,
}

\section{Introduction} \label{introduction}
Autonomous driving technology heavily relies on a combination of cameras and LiDAR sensors, mostly due to the complementary benefits that LiDAR sensors bring to most detection pipelines. LiDAR sensors provide dense and detailed point cloud maps using rotating sensors with spherical/semi-spherical coverage of the surrounding area. These sensors' pre-processed data features can be easily integrated with images from camera sensors in autonomous driving.
However, LiDAR sensors have shorter wavelengths, causing the following limitations in LiDAR object detectors \cite{9000872}. (i) Signals are highly prone to errors under poor weather conditions and occlusion (ii) they have a relatively shorter sensing range.  

 In contrast to LiDAR signals, radar’s Frequency Modulated Continuous Wave (FMCW) signals operate at the millimeter-wave (mmW) band, or in the frequency band between 30 to 300 GHz. The mmW band is much lower than visible light, which allows radar signals to go through occlusion particles such as smoke and dust, enabling radars to function more robustly in extreme weather conditions. Furthermore, radar signals' longer wavelength (mmW) provides a larger range for detection with acquisition capabilities reaching up to 3000 meters. Radars are also more accessible and cheaper to introduce to dynamic systems compared to LiDARs \cite{radical}.

\begin{figure}
\includegraphics[width=\textwidth]{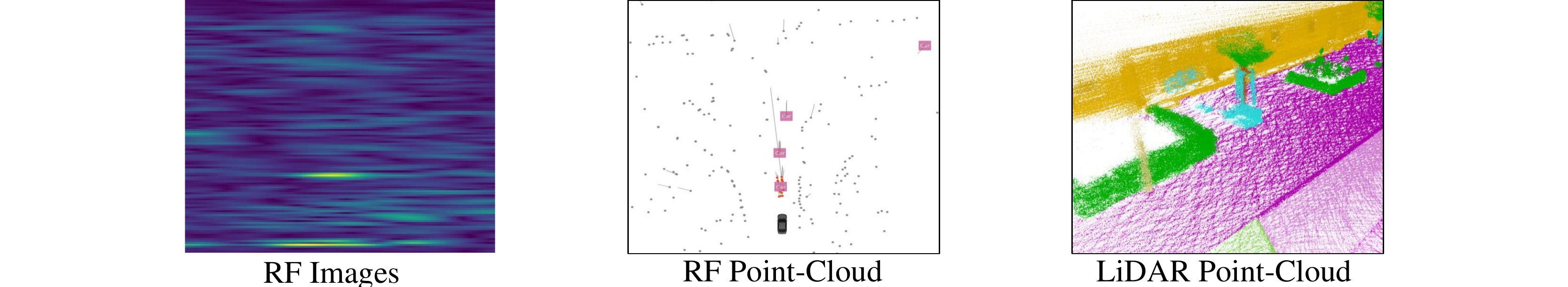}
\includegraphics[width=\textwidth]{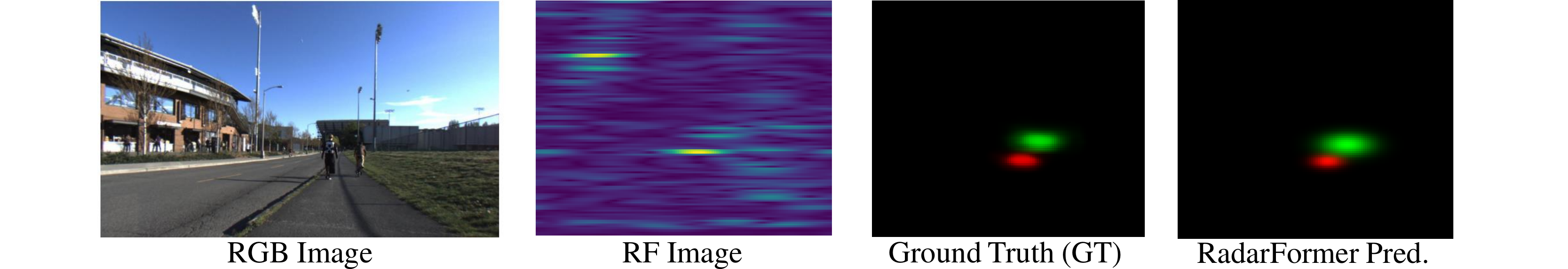}
\caption{The first row shows an example of different data samples from a radar-frequency image-like data point (from \cite{rodnet}),
a point-cloud radar-frequency data sample (from \cite{radar_scenes_dataset}), and a point-cloud LiDAR sample (from \cite{kitti}).
The RF image heatmap signifies the magnitude of the echoed radar signal from the radar, with blue being the minimum. The second row shows a sample RGB image and its corresponding RF image. Our model, RadarFormer, takes in the RF image only and produces a heatmap prediction with the object class (shown in different colors), illustrated in the rightmost image.}  \label{fig:introduction_fig}
\end{figure}

It is possible to extract point cloud data from raw radar signals, however, it is more common to extract them as radar frequency (RF) image-like data. Radar signals are sent and received through a multi-input multi-output (MIMO) antenna array, which is then passed through a series of fast Fourier Transform (FFT) to extract range-angle-doppler maps. These maps compactly describe the 3D space in the range plane (distance to detection), azimuth plane (angle of arrival), and doppler information (relative velocity) \cite{10.3389/fnins.2022.851774}. An example of RF data and LiDAR data in comparison can be seen in Figure \ref{fig:introduction_fig}. 
In the RF image, we see an RF range-azimuth (RA) map shown in an image-like processed format. The second image shows RF point cloud data, which can be compared to the LiDAR point cloud data, where we see that LiDAR sensors provide more detailed object descriptions compared to radar data. However, radar data can provide readings from a longer range and contain velocity information, as discussed earlier. 
Multiple works demonstrated the use of radar data as a feasible alternative to cameras and LiDARs in object classification \cite{autom-radar-class,vehicle-class,radar-id}.  
Recently, \cite{radsegnet,rodnet} explored radars as an opportunity to be fused with other sensors such as camera-radar fusion to produce more accurate predictions. 

The increasing availability of radar frequency data now has opened the path to explore more complex approaches for radar perception. Common radar datasets provide a variety of RF data as input, for instance, \cite{rodnet} provides only the RA map, while \cite{carrada} provides RA, Range-Doppler (RD), and Range-Azimuth-Doppler (RAD) maps. Some datasets \cite{raddet,carrada} provide the original radar tensors in addition to the maps, while others \cite{radical} provide the digitized output after the Analog to Digital Converter (ADC) stage. Radar data can also be provided as point cloud data, as was shown earlier and provided by \cite{radar_scenes_dataset} with range-azimuth information in a 3D space point cloud data. The previous works for radar object detection used computationally expensive models that use large 3D convolutions and might be impractical in generating rapid real-time predictions. 

In this work, we propose a transformer-based deep-learning model that operates on radar frequency data exclusively and produces state-of-the-art results in object detection and classification. The proposed model is lightweight in size and generates inferences in real-time, making it suitable for the task of autonomous detection. Figure \ref{fig:introduction_fig} (bottom row) shows a camera RGB image, a radar RF image, and the corresponding ground truth annotation from CRUW dataset \cite{rodnet}. The proposed RadarFormer takes only the RF image as input (without the RGB image) and produces a heatmap prediction of the localized object class shown in the rightmost image. The key contributions of the proposed approach are:
\begin{itemize}
\vspace{-2mm}
    \item We explore the effectiveness of vision transformers in radar perception and introduce a novel architecture, RadarFormer, for real-time radar object detection. To the best of our knowledge, we are the first to introduce a transformer-based architecture for RF maps data for the task of object detection.  
    \item We propose a channel-chirp-time merging module that contributes to reducing the size of radar perception models and using less computationally expensive modules.
    \item Our proposed method, RadarFormer, achieves state-of-the-art performance with one-tenth the model size of the previous state-of-the-art model 
    and a two-times faster inference speed. 
    
\end{itemize}

\vspace{-5mm}

\section{Object Detection on Radar Data} \label{radar object detection}
Radar data is usually visualized as two-dimensional or three-dimensional maps with two sets of channel dimensions. The first is the real and complex part of the RF signal, and the other is the chirps of the echoed signal. This data form makes them suitable for multiple deep learning models utilizing Convolutional Neural Networks (CNN). While conventional CNNs do not work with point-cloud data, multiple models were still explored through the voxelization of input data, such as LidarMultiNet \cite{lidarMN}, PanopticPolarNet \cite{panopticpolarnet}, or works that encourage interaction between the model CNNs and voxelized input like JS3C-Net \cite{sparse-single}. For 2D image-like data, multiple works detail the process of generating RA, RD, RAD, and RAMaps \cite{carrada,10.3389/fnins.2022.851774,rodnet}. The output of said maps is then passed through a Constant False Alarm Rate (CFAR) algorithm, which checks the amplitude of pixels to determine their magnitude relative to the average noise level in surrounding pixels, and classifies pixels as `object' and `non-object'. After CFAR, object classes are then determined following different techniques. For example, CRUW data set uses a 3D object localization and class recognition to generate ground truth for the data \cite{monocular}. It is common to also use temporally and spatially aligned cameras and LiDars to generate annotations for object detection purposes without classifications \cite{radical}. In a similar fashion, the work in \cite{radarfusion2019} generated the annotations using radar LiDAR fusion with the latter being the ground truth. Other works \cite{carrada} use clustering techniques that rely on the rich doppler information properties of radars to create clustered annotations, with some manual annotations as a quality check \cite{radar_scenes_dataset}.

Following the generation of data sets, most of the works in the literature heavily rely on CNN-based deep learning models. RADDet \cite{raddet} uses a radar-tailored ResNet backbone followed by a YOLO-inspired \cite{yolo} dual detection head to produce object detections and classifications. Encoder-decoder style models are very popular and were adapted differently to different datasets. RODNet \cite{rodnet} uses a stacked-hourglass model to generate predictions, while TMVA-Net \cite{carrada_model} uses a temporal-multi-map encoder decoder in their CARRADA data set. Other works like RadSegNet \cite{radsegnet} and LidarMultiNet \cite{lidarMN} introduces the encoder-decoder block after the input voxelization step to point cloud data. Following this discussion, we notice a pattern in the over-reliance on CNN-based deep learning models. 
While they perform adequately in most detection-based models, our proposed model extends beyond CNNs to include more developed deep learning techniques for radar perception systems.

\begin{figure}
\centering
\includegraphics[width=.9\textwidth]{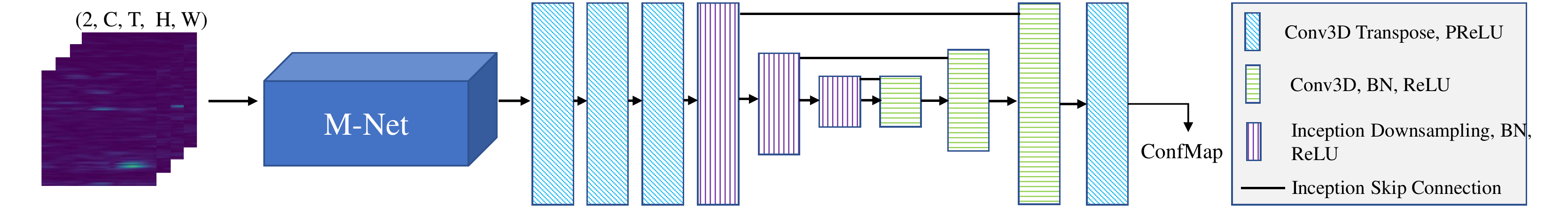}
\caption{Original RODNet hourglass with inception model as per \cite{rodnet}.} \label{fig:rodnet}
\end{figure}

Given the benefits of radar frequency data for object detection and classification, multiple radar datasets have been collected for this purpose. The Camera-Radar of University of Washington (CRUW) data set \cite{rodnet} contains RF images collected through radar sensors with synchronized cameras connected. The FMCW mmW radars along with the camera collect synchronized radar maps and images at 30 frames per second (FPS) with 255 chirps per frame and a range and azimuth resolutions of 0.23 m and 15\degree, respectively. The data set contains three classes: pedestrians, cyclists, and cars, with around 5 objects per frame on average. RODNet was proposed as a `student' module that learns alongside a camera-radar fusion (CRF) cross-modal approach. RODNet alone takes only the RF images and produces confidence maps (ConfMaps) which are passed later into a location-based non-maximum suppression (L-NMS). 
Fig. \ref{fig:rodnet} illustrates the RODNet model architecture, which consists of a chirp merging module (M-Net) that downsamples chirps into one layer, followed by the stacked-hourglass architecture featuring the temporal inception convolutional layers. The model reported an average precision (AP) and average recall (AR) of 77.40\% and 80.80\%, respectively, without the CRF and using only a camera-only supervision method, which will function as our baseline in terms of the accuracy of predictions.

\begin{figure}
\includegraphics[width=\textwidth]{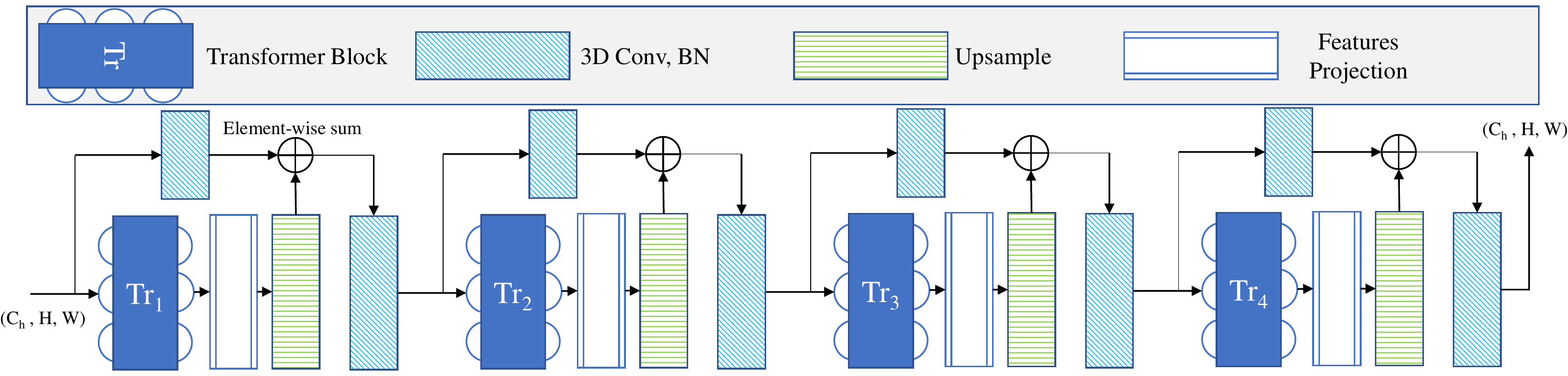}
\includegraphics[width=0.95\linewidth]{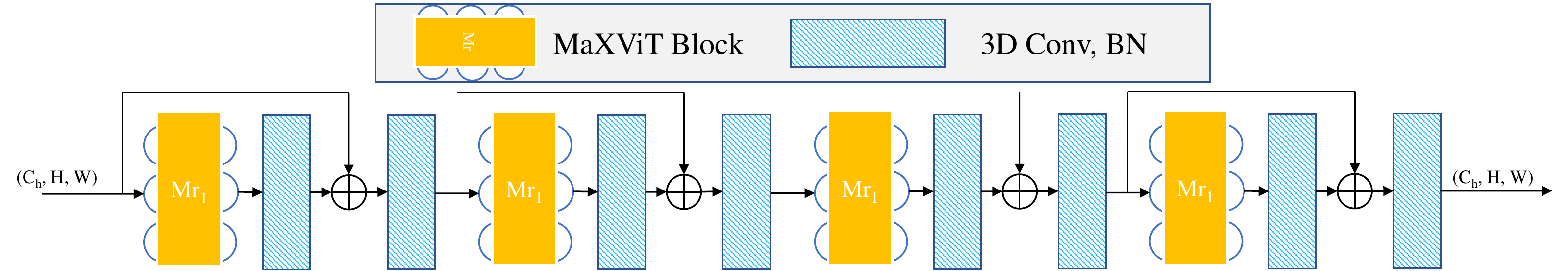}
\caption{Our transformer-based proposed models, 2D transformers (top) and RadarFormer (bottom). The input to both models is the output from the output of the channel-chirp-time merging module, which will be discussed in Section \ref{2d discussion}. The output of the ViT transformers downsamples the input, requiring an upsample block to retrieve the original resolution. 
The same flow is followed in the MaXViT-based model, without any resolution changes to the inputs/outputs.} \label{fig:radarformer}
\end{figure}

\vspace{-1mm}

\section{The Proposed RadarFormer} \label{proposed model}
While CNNs have been a dominant architectural design block for a lot of tasks in both image recognition and radar-based recognition \cite{7780459,7298594}, we propose a transformer-based architecture that utilizes recent developments in deep learning transformer techniques, shown in Fig. \ref{fig:radarformer}. Our model introduces a hybrid model between CNNs, transformers, and multi-axis attention following the work done in \cite{maxvit}. We also introduce an updated channel-chirp merging module that includes the temporal domain. Our module extends the merging of extra channels to include residuals connecting the temporal domain in the downsampling/upsampling stream and is shown in Fig. \ref{fig:mnetanddownsampling}.

\vspace{-2mm}

\subsection{Transformers} \label{transformers}

In recent years, vision transformers (ViTs) \cite{vit} were introduced as a new paradigm that is fully free of convolutional networks and produced state-of-the-art results for image recognition tasks. Various approaches and versions of self-attention modules were proposed with some works reporting state-of-the-art results in object detection without the use of CNNs \cite{swin1,swin2}. Transformers were originally developed for Natural Language Processing (NLP) tasks using 1D sequence inputs. Transformers were then repurposed for images in ViTs, taking in a 2D input image of size $x \in \R^{H\times W\times C}$, where $(H,W)$ is the image resolution and $C$ is the number of channels. Said images are then flattened into a sequence of patches, $x_p \in \R^{N\times (P^2\cdot C)}$, where $x_p$ denotes a single patch, $N = \frac{H\times W}{P^2}$ denotes the length of said sequence, and $(P,P)$ denotes the size of every patch. Transformers were expanded to 3D data by including temporal information, volumetric medical images, or spatial 3D images \cite{lahoud,detr3d,vtunet,9438625}. In 3D sequences, we have an input of size $x \in \R^{D\times H\times W\times C}$, where $D$ is the depth of the data (or the third domain of the respective application). Here, we have a sequence of patches shaped as $x_p \in \R^{N\times (P^3\cdot C)}$ where $N = \frac{H\times W\times D}{P^3}$ and $(P,P,P)$ is the size of the patch. These patches are associated with positional embeddings to preserve the positional information tied to the original data and then passed into the transformer encoder. The basic transformer encoder in ViT consists of a multi-head self-attention (MSA) and multi-layer perceptron (MLP) blocks, which produce an output $x_z \in \R^{N\times S}$, where $S$ is the projection following the MLP blocks. The MSA block learns a mapping between a query (q), a corresponding key (k), and a value (v) representation of the encoder output, measured by $\text{[\textbf{q,k,v}]} = x_z\textbf{U\textsubscript{qkv}}$
where $U\textsubscript{qkv}$ is the projection MLP weights. The \textbf{q} and \textbf{k} representations are then used to find the attention weights $A$ through
\begin{equation}
\textbf{A} = \text{Softmax}(\frac{\textbf{qk}^\top}{\sqrt{S_l}})
\end{equation}
$S_l$ a scaled version of $S$ by a factor $l$ and is set to $\frac{S}{l}$, keeping the total number of parameters constant given a variation in the number of key values $\textbf{k}$. Using the attention weights, we then measure the self-attention $SA$ with
$\text{SA(}x_z\text{)} = A\text{v}$.
MSA scales the previous expression up to a sequence of self-attention heads in the form of a vector with each head having its own unique set of weights in $U\textsubscript{msa}$, described by

\begin{equation}
\text{MSA(}x_z\text{)} = \text{[SA\textsubscript{1}(\textbf{x\textsubscript{z}});SA\textsubscript{2}(\textbf{x\textsubscript{z}}); ... ;SA\textsubscript{m}(\textbf{x\textsubscript{z}})]}\textbf{U\textsubscript{msa}}
\end{equation}

The data set we operate on can be adjusted to work in both 3D and 2D (including the temporal domain or with a downsampled temporal domain), as will be shown in Section \ref{2d discussion}. RadarFormer's early stages used expensive 3D transformers that are based on ViT \cite{vit} and improved by \cite{unetr}. While this provided good results, the computational complexity was still high. We can use 2D transformers by passing our input data through our proposed channel-chirp-time merging stream (Section \ref{2d discussion}), providing a much lighter model and higher accuracy as well. Using 2D data allows us to explore a computationally inexpensive 2D variation of attention as we will see in Section \ref{attention}.

\vspace{-2mm}

\subsection{Attention Variation} \label{attention}
Following ViT, multiple works have been developed that aim at more robust vision transformer models.
Some works introduced many features to image transformers such as hybrid models between transformers and CNNs \cite{hybrid1,hybrid2,hybrid3}, with many surveys evaluating and contrasting these models and their variations \cite{survey2}. 
In addition to introducing hybrid models, variations in attention modules were also explored by many works \cite{survey3}. Our module uses a transformer block, called MaXViT (Multi-Axis Vision Transformer) \cite{maxvit}, which consists of an inverted residual block, MBConv \cite{mbconv}, followed by an attention block and grid attention. The MBConv consists of 3 convolutional layers with a wide-narrow-wide style of channels, and a small-large-small style of kernel sizes, with a residual connecting the first convolution to the last one. The multi-axis attention block creates windowed partitions of the input and performs self-attention on these partitions, whose shape is $(\frac{H}{P},\frac{W}{P},P\times P, C)$, creating non-overlapping windows of size $P\times P$ following the notation used earlier. Similarly, a grid partitioning module uses a $G\times G$ uniform grid to partition the input with adaptive size $\frac{H}{G}\times \frac{W}{G}$, resulting in a dilated mix of tokens that provide global information. In the module proposition, stacking both window and grid attention provides local and global contexts in transformer operations, hence the name multi-axis attention. 

As an attempt to improve the transformers models, we explored multiple variations of transformer models and attentions along with convolutional combinations. The first is a high-low attention (HiLo) \cite{hilo} approach that splits attention into a high-resolution branch and a low-resolution (downsampled) branch. This approach did provide more consistency in training but did not perform adequately, where it capped at 74.2\% AP. This led us to explore the attention and resolution variation of the models more thoroughly. To this end, we explored a high-resolution transformer (HRFormer)-like architecture, following the work proposed in \cite{hrformer}. HRFormer was very computationally expensive both in its 3D and 2D variants and did not perform well. The number of trainable parameters, 872.9 million, was too large to justify its use, and too large for the data set to train and performed poorly as expected. In our proposition, MaxVIT blocks performed consistently better with a lot of variations when compared to other baseline models. Other architectures like UNETR \cite{unetr} and UNETR++ \cite{UNETR++} were computationally expensive in their 3D format and did not provide an accuracy high enough to justify their use. However, downsampling to 2D and using a UNETR/ViT-inspired transformer design provided a good baseline for transformer architectures which we referred to as '2D Transformer' in Fig. \ref{fig:radarformer}. We discuss the quantitative results of relevant models in Section \ref{quantitative}.

\subsection{Use of 2D Information} \label{2d discussion}
\begin{figure}
\includegraphics[width=\textwidth]{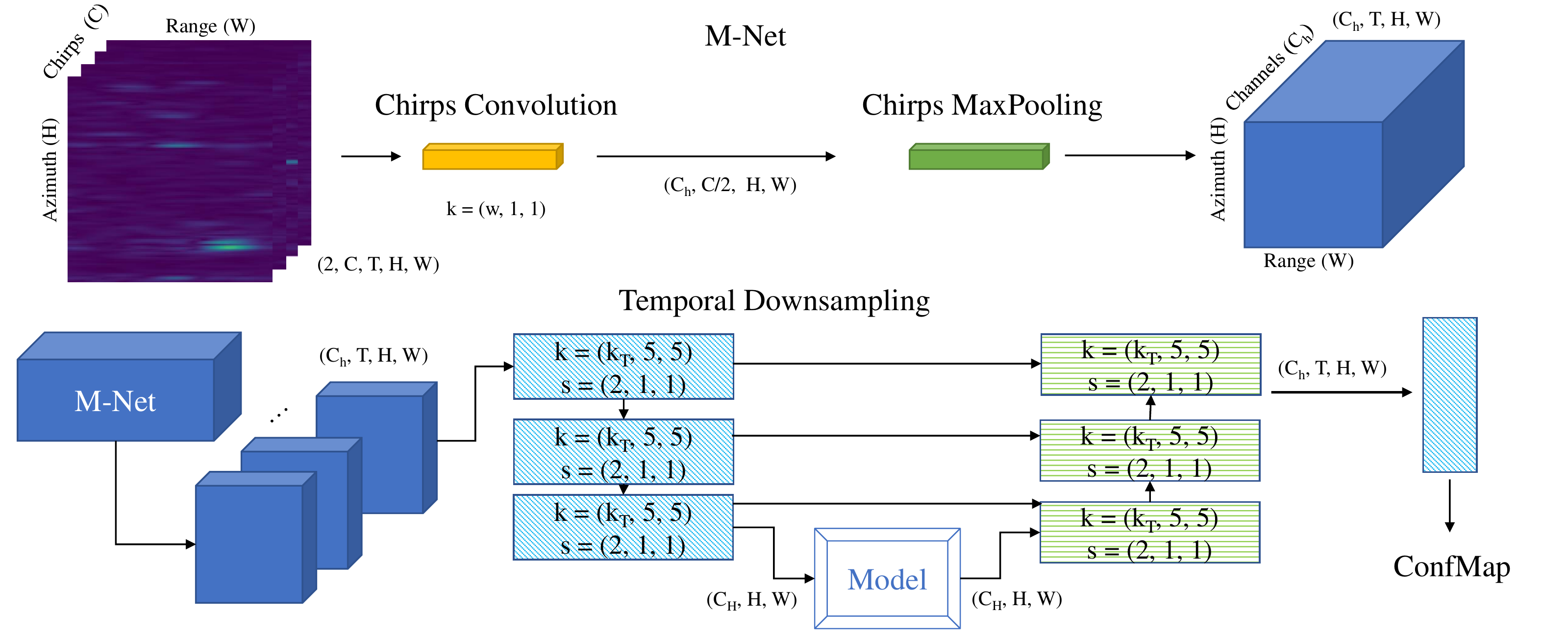}

\caption{The M-Net module (top) reported in the original RODNet architecture and used by us and proposed temporal downsampling/upsampling flow (bottom). The M-Net module takes in the full RF image with shape $(2, T, C, H, W)$ and merges the channels (2) and the chirps $(C)$ into one dimension, $C_h$. The second module downsamples the temporal domain to receive a 2D tensor of shape $(C_h, H, W)$. The tensors are connected via element-wise addition to their upsampling counterpart after the models forward the input. Both modules are cascaded and output a 3D output of shape $(C_h, T, H, W)$.}\label{fig:mnetanddownsampling}
\end{figure}

In the original model of RODNet, the input has a shape of $(B, 2, T, C, H, W)$, where $B$ is the batch size, $C$ is the number of chirps, $T$ is the window size (temporal), $2$ is the number of RF channels (being a constant referring to the real and imaginary magnitudes of RF signals), and $H,W$ are the height and width of the RF image, respectively. The RF channels and chirps are merged into one dimension using M-Net, a channel-chirp merging module suggested in \cite{rodnet}. We extend M-Net to go beyond channel-chirp merging and propose a module that includes temporal merging as well whose inputs and outputs are shown in Fig. \ref{fig:mnetanddownsampling}. The chirps and RF channels dimensions are compressed first into one channel with dimension $C_h$, shown as $(B, C_h, T, H, W)$, and then downsampled in the temporal domain to be passed into the models. However, we preserve temporal context through a temporal residual connection towards the upsampling stream at the end before generating the ConfMaps. 

The motive for downsampling comes from two observations.
First, 3D convolutions and temporal 3D convolutions are extremely computationally expensive and their weights have a large size, especially when using many filters. This is undesirable in the context of autonomous vehicles that have controllers with limited computing and storage capabilities. Second, we noticed that downsampling 3D inputs into 2D, performing all deep learning operations after downsampling the frames into one frame, and upsampling before inference with residuals did not compromise the accuracy to a noticeable level. The downsampling however reduced the inference time to one-fifth of the original value and the size of the weights to one-tenth of the original value. We can conclude that the effect of the time domain is less pronounced when treated as separate channels and employing expensive 3D convolutions or 3D transformer modules like UNETR is not necessary. Other models in radar perception used temporal downsampling techniques like TMVA-NET \cite{carrada_model}. The reduced base RODNet which has a temporal downsampling/upsampling stream added to is referred to as '2D CNN' from this point forward. It is worth mentioning that downsampling is more useful than distributing channels. We noticed that when we skip the 3D downsampling module and instead stack our model input to the shape $(T\times B, C_h, H,W)$ we lose a lot of contextual temporal information. This information was retained through the convolutional filters and the residual connections between the input and output. 

We noticed that RODNet's performance degraded by a negligible margin by downsampling the temporal domain to a single channel and performing most of the operation with reduced numbers of channels with 2D convolutions. In addition, the model did produce heavily fluctuating results as the model trains, which initially was attributed to the configuration of the learning rate, however, altering the learning rate value, the scheduling, and the optimizer \cite{sgdr} did not make the model converge to a consistent value. Furthermore, the reproducibility of networks trained on this data set was inconsistent and sometimes inaccurate. The same model could be trained using the same configurations and yield completely different prediction accuracies. Replicating the results of reported values of RODNet was also not feasible despite attempting to train the same model multiple times. Introducing residual connections between the input downsampling convolutions and the output upsample convolutions helped remedy this issue. The difficulty of learning consistent weight sets is one of the challenging aspects of this data set and application, but we noticed better reproducibility of transformer-based architectures compared to the CNN-only approaches. 

\vspace{-2mm}

\subsection{Effect of Receptive Fields and Residuals} \label{receptive field}

Transformers did introduce a large reduction in required computational complexity at the early stages of our model design, providing a large margin for us to utilize more complex architectures and modules. To take advantage of both transformers and CNNs, we noticed that introducing CNNs before and after the transformers along with residual connections following our design in Fig. \ref{fig:radarformer} improved the model's accuracy. This was in line with the suggested conclusions of \cite{earlyconv}, which was encouraging to explore two main ideas in our experiments. The first is exploring the effects of varying the number of convolution layers, input/output channels, and residual connections before and after the transformers. The second is to assess the effect of varying the receptive fields and the size of the convolutional layers' kernels. This discussion excludes the first transformer layer of both models due to it being preceded by 3 convolutional downsampling layers. For the first point, using two convolutional layers before the transformer and one after the network provided the best results without residuals. Introducing residuals with element-wise addition and a convolutional layer that equalizes the number of channels, followed by a convolution after the element-wise addition, provided similar performance with a slight decrement in required computational time. We noticed that MaXViTs perform the best without downsampling and the convolution in the residual connection is not necessary and does not improve the performance either with MaXViTs, providing a lighter model in this aspect by removing it. For the second point, varying the kernel sizes of said layers did contribute to better learning of data. Increasing the size of the kernel as we go deeper in the network allows for learning of a more general receptive field that gets improved by the transformer's global/local attention dynamic. In a similar fashion, establishing residual connections between the input 3D downsampling to the output 3D upsampling compensates for the information loss in propagating through the network and improves the upsampling stream.

\vspace{-3mm}
\subsubsection{Implementation Details} We train our models on a single NVIDIA A100 GPU with Adam optimizer and an initial learning rate of $10^{-4}$ with step decay.

\section{Experiments} \label{experiment}
\subsubsection{Dataset Details} \label{dataset}

We build and test our model on the CRUW radar data set \cite{rodnet}. CRUW consists of roughly 400k frames of recorded driving sequences. The data is processed to be represented as Range-Azimuth Heatmaps (RAMap)s, which describe a bird's-eye view of the scene seen from the ego-vehicle. The $x$-axis depicts the azimuth plane, describing the angle, and the $y$-axis depicts the range plane, describing the distance to the object, with the intensity describing the magnitude of the RF signal. These can be described as an image with a resolution of 128$\times$128 each, with a sample shown in Fig. \ref{fig:introduction_fig}. The acquisition setup consists of two 77GHz FMCW antennas that collect 256 chirps every frame (30 frames a second), and only 4 are chosen out of these chirps (0, 64, 128, 192). The details of the data we use and inputs will strictly follow the work done in \cite{rodnet} which most of our work will use as an evaluation baseline. This work will also focus on the behavior of deep learning modules with radar-based data, which has not been studied extensively in most works regarding radar data. Unlike the model associated with the CRUW data set, RODNet, our aim was to create a deep learning model that uses radar data without any sort of sensor infusion, which was illustrated in the model discussed in Section \ref{proposed model}. The input to the model is a $(B, C_h, H, W)$ input tensor, with the batch size $B$, the number of channels $C_h$, and $(H,W)$ is the resolution of the RF image. We note that this is the output of the downsampling/upsampling module discussed earlier in Section \ref{2d discussion}. The RF image resolution $(H,W)$ is fixed at $128$. There are 40 sequences reserved for training and 10 sequences for testing. Testing annotations and images are not publicly shared and evaluation of the testing is done on a private evaluation server for the RODNet2021 challenge \cite{challenge}.

\subsubsection{Evaluation Metrics} \label{loss}

We use the same evaluation metrics as in \cite{rodnet} throughout all of our evaluations. RODNet uses an object location similarity (OLS) metric that takes the role of Intersection over Union (IoU). 
OLS is then passed to a location-based non-maximum suppression (L-NMS) algorithm to generate confidence maps (ConfMaps). ConfMaps in the range-azimuth represent predicted locations for the objects, with multiple channels attributing the location to a class. Similar to previous work for pose estimation \cite{stackedhourglass}, the output is a gaussian heatmap-like prediction with a mean equal to object location and variance attributed to the object class and scale information. Our main evaluation metric will be the average precision (AP) and average recall (AR) calculated through the variation of OLS threshold between 0.5 to 0.9 with steps of 0.05.

\subsection{Baselines} \label{baseline}
The baseline to the CRUW data set is the RODNet model associated with it, reporting an AP of 77.40\% and AR of 80.80\% using camera-only (CO) supervision. Instead of using the reported inference times, we instead retrain the model and report inference speed using our equipment to provide a consistent and scalable assessment and comparison. The model reports 61.2 million total trainable parameters (single stack hourglass). The time it takes for a single backpropagation iteration with a window size of 16 for this model on our setup is 1920 ms, and the average inference time is 148 ms. This is associated with 47664 GMAC ($\times 10^{9}$ multiply-accumulate) operations. The listed parameters will be the main comparison points to our model.

\subsection{Quantitative Results} \label{quantitative}

Due to the recency of emerging RF-based data, and the lack of other works to compare to, we compare our proposed models, 2D transformer and RadarFormer, to RODNet considering the latter the state-of-the-art method. Table \ref{tab:ap_comp} shows the performance comparisons of mentioned models. The evaluation metric here is AP and AR, each is split into four categories: $PL$, $CR$, $CS$, and $HW$, referring to parking lot, campus road, city street, and highway data categories, respectively. Both AP and AR are then averaged into $AP_{total}$ and $AR_{total}$. We can see that RadarFormer achieved an AP almost on par with RODNet, with a higher AR. They varied in their performance in different situations, where we noticed that RadarFormer had slightly better performance in the parking lot scenarios, while RODNet performed noticeably better on campus roads. In city streets, RadarFormer had a higher AR than RODNet, but a lower AP, and on highways, RadarFormer outperformed RODNet significantly. Looking at total AP, the models perform comparably the same, but RadarFormer had a tendency to have a higher recall, on average, implying that RadarFormer has a tendency to produce fewer predictions, but an inclination to have higher confidence for the produced predictions. We also note that 2D CNN, which is a 2D version of RODNet that utilized our channel-chirp-time merging module did not compromise the accuracy greatly when compared to the retrained version of RODNet. However, we will see in Section \ref{model comparisons} how this module reduced the computational cost and size of RODNet to a large degree.

\begin{table}
\centering
\caption{Quantitative comparisons of AP and AR on the CRUW data set with CO supervision, categories are described in Section \ref{quantitative}. RadarFormer performed comparable to RODNet, being the accessible and replicable state-of-the-art model for this data set. The exact values of RODNet were not replicable, so we use the values reported in the RODNet challenge server \cite{challenge} and report the values trained of the publicly available model (\textbf{*}). We discuss the discrepancy in the results in Section \ref{2d discussion}. 
}\label{tab:ap_comp}
\begin{tabular}{|l |c |c |c |c|c|c|c|c|c|c|}
\hline
\multirow{2}{*}{Method} & \multicolumn{2}{c|}{Total} & \multicolumn{2}{c|}{PL} & \multicolumn{2}{c|}{CR} & \multicolumn{2}{c|}{CS} & \multicolumn{2}{c|}{HW} \\ 
& \multicolumn{1}{c}{AP} & \multicolumn{1}{c|}{AR} & \multicolumn{1}{c}{AP} & \multicolumn{1}{c|}{AR} & \multicolumn{1}{c}{AP} & \multicolumn{1}{c|}{AR} &  \multicolumn{1}{c}{AP} & \multicolumn{1}{c|}{AR} &  \multicolumn{1}{c}{AP} & \multicolumn{1}{c|}{AR}\\
\hline
RODNet\textbf{*} & 72.32 & 79.62 & 94.52 & 95.59 & 68.12 & 72.50 & 52.30 & 72.13 & 67.48 & 71.63 \\
\hline
RODNet \cite{challenge} & \textbf{77.40} & 80.80 & 95.50 & 96.40 & \textbf{75.30} & \textbf{78.40} & \textbf{66.10} & 71.70 & 68.00 & 72.20 \\
\hline
2D CNN & 71.58  & 81.52 & 94.93 & 96.04 & 66.52 & 73.56 & 52.15 & 75.62 & 69.23 & 74.13 \\
\hline
2D Transformer & 75.03 & 81.99 & \textbf{96.68} & \textbf{97.55} & 68.51 & 77.04 & 58.67 & 75.08 & 70.16 & 72.46 \\
\hline
RadarFormer & 77.18 & \textbf{83.45} & 95.88 & 96.99 & 71.74 & 77.66 & 61.19 & \textbf{76.46} & \textbf{74.37} & \textbf{77.30} \\ 

\hline
\end{tabular}
\end{table}
\vspace{-5mm}

\subsubsection{Model Comparisons} \label{model comparisons}

We further compare the models' computational cost and inference times using Table \ref{tab:models_comp}. We note that we use MAC operations instead of floating-point operations (FLOPs) due to PyTorch's inclination to use MAC operations. The M-Net module used earlier remains constant and its performance is reported separately. The word `size' is used interchangeably with the number of parameters. Despite RODNet having reported a marginally higher accuracy in CO supervision predictions, the model requires almost twenty times more MAC operations than our proposed model, while also being ten times as big in size. This is significant in regards to employing said models on devices that don't have much computing power. Similarly in the same table, the inference and backpropagation (BP) times are aligned with the MAC discussion, where RadarFormer does provide inferences at roughly half the time of RODNet. We would like to note that these numbers are relative, and they scale up and down based on the used GPU for training and inference. Furthermore, we normalize and take into account the used window sizes and batch sizes for the training and testing. We also point out that reporting the BP time for a single iteration instead of training time per epoch is a more accurate measurement for this data set to factor out the loading time from the storage devices to the CUDA GPU. 2D CNN is a very lightweight and fast model that compromises the accuracy marginally compared to RODNet, while 2D Transformers provides a middle-ground between training and inference time and accuracy, but at the cost of having a large model size and the number of parameters.

\begin{table}
\centering
\caption{Comparisons of ($10^9$) multiply-accumulate (GMAC) operations, the number of parameters (No. Parm.) in millions (m), back-propagation (BP) time per iteration, and inference times between the different models (inclusive of M-Net). All time units are in milliseconds (ms). M-Net has no stand-alone inference or BP time since all models use it. The downsampling/upsampling streams are counted in our proposed models. All predictions and inferences were adjusted and normalized to take into account the batch size, window size, and test stride for equal comparisons.}\label{tab:models_comp}
\begin{tabular}{|l|c|c|c|c|}
\hline
Model &  GMACs & No. Parm. (m) & BP Time (ms) & Infer. Time (ms)\\
\hline
M-Net & 0.805 & 0.224 & - & - \\
\hline
RODNet \cite{rodnet} & 47664 & 61.22 & 1920 & 148.17\\
\hline
2D CNN & 388.6 & 2.71 & 330 & 51.92 \\
\hline
2D Transformer (ours) & 1950 & 20.88 & 380 & 60.38 \\
\hline
RadarFormer (ours) & 2123 & 6.42  & 700 & 84.35 \\
\hline
\end{tabular}
\end{table}

\vspace{-7mm}

\subsection{Qualitative Results} \label{qualitative}
\begin{figure}
\includegraphics[width=\textwidth]{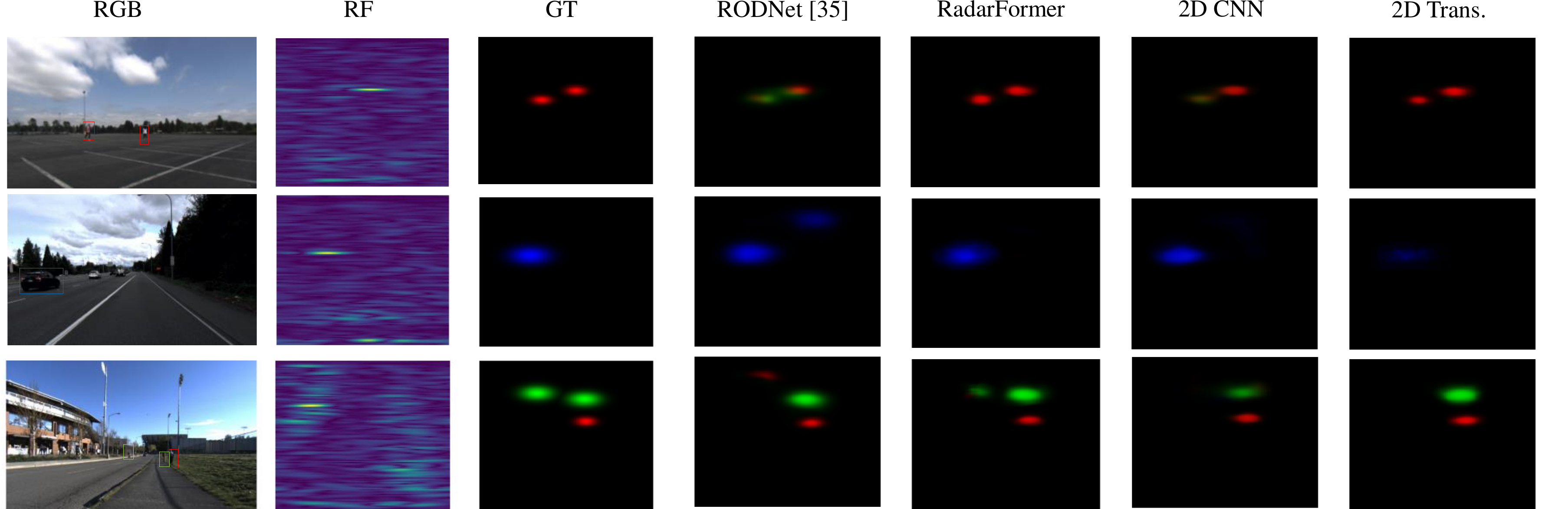}
\caption{Test cases for all models mentioned. Pedestrians, cyclists, and cars are referred to with red, green, and blue, respectively in the prediction and ground truth map. We see that RadarFormer provides predictions with acceptably high confidence when they are presented, in contrast to other models that are prone to occasional false labeling.} \label{fig:qualitative_full}
\end{figure}

We refer to Fig. \ref{fig:qualitative_full} for samples of the predictions from the mentioned models. It shows the original RGB image, RF image, ground truth (GT), then the prediction heatmap of the models for three samples. We also note that since we do not have access to the images of the test set, the models were re-trained on a 90\% split of the training data, and these predictions were generated on the unseen 10\% of the data. As discussed earlier and is evident by the relatively high AR of RadarFormer, the model produces predictions with relatively higher confidence when it decides on an object and a class, depicted through the accurate class label and higher brightness of the heatmap. We also notice that RadarFormer performs better in scenes with far objects as can be seen in the third row.

\begin{figure}
\includegraphics[width=\textwidth]{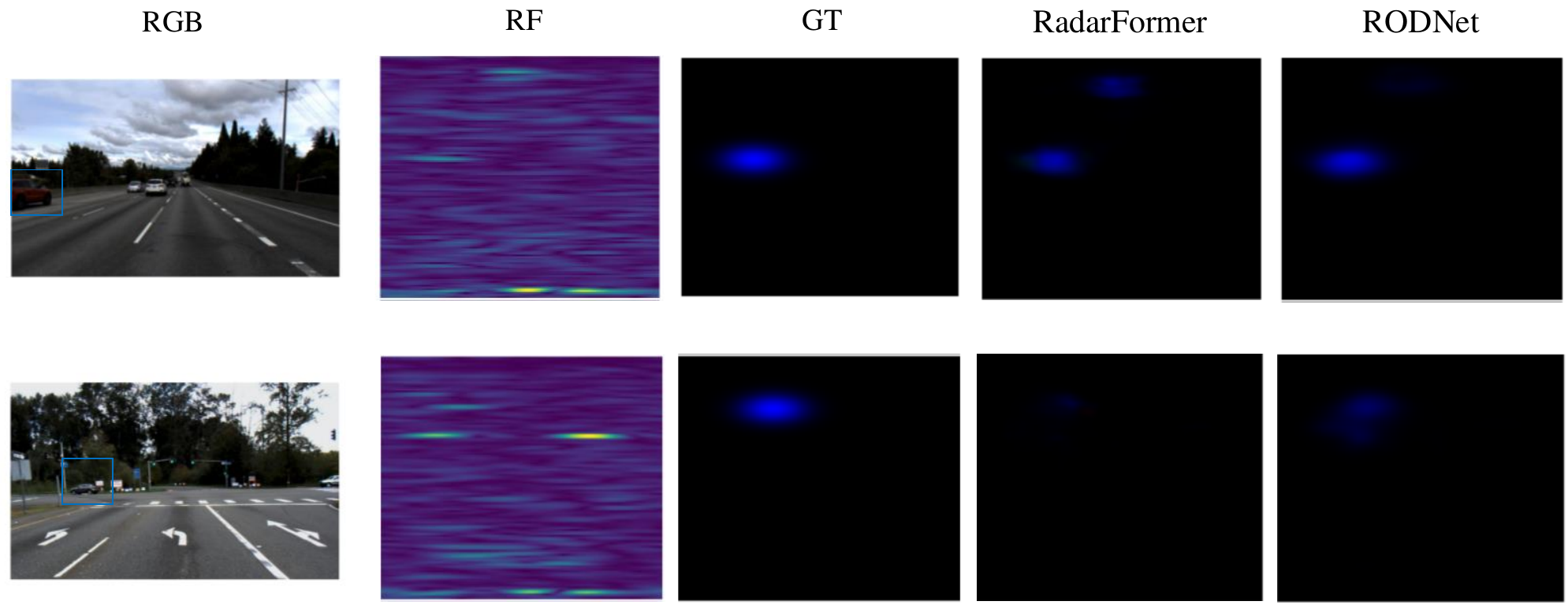}
\caption{Incorrect predictions generated by our proposed model, compared with predictions from RODNet. We note that despite there inaccurate annotations cases like this, the model attempts to predict objects that were not annotated. We can see this evident in the top row where we see predictable objects in the RGB images and the RF images, but not present in the ground truth. The second row is a difficult case where RODNet generated a better prediction than RadarFormer.} \label{fig:fail_cases}
\end{figure}

We further explore the incorrect prediction cases of RadarFormer and list some of the model's limitations. Failure is most common due to difficult cases or due to inaccurate annotation of the ground truth, and we show examples of this in Fig. \ref{fig:fail_cases}. First, the model tends to generate predictions of objects that it perceives in the RF images but are not present in the ground truth, which can be seen in the first row. We can see a second car, however, the ground truth doesn't show its presence and is then determined as an incorrect prediction case. We also notice that the model does not generate predictions of objects that are close to other objects within sensor's view, as can be seen in the second row.

\vspace{-1mm}

\vspace{-3mm}

\subsection{Ablation Studies} \label{ablation}
\vspace{-1mm}
\textbf{Data Format and Model Variations} We use a window size of $32$ ($32$ frames) in the reported results, except for the RODNet baseline which was configured to 16. Using different window sizes between 8, 16, 24, and 48 did not yield any noticeable improvement to the model and reported degraded performance (1 to 2\% less than using 32). This was evident when operating with and without temporal downsampling. Similarly, we varied the patch size of the transformers, the window size of MaXViT, and the MLP sizes. The BP and inference times remain relatively constant, however, the required video ram (VRAM) increases exponentially with larger attention windows without any improvement to the accuracy. Similarly, the size of the window attention for MaXViT should remain a fraction of the image resolution (window size of 7 for an RF image of resolution 128). Using the base window size of 7 yielded consistently higher AP than other counterparts. Increasing the number of neurons in transformer MLP caused the model to not converge to a high enough accuracy compared with a lower number of neurons, so we use an MLP ratio range between 20 and 150, depending on how deep the model goes (smaller ratio for deeper models). Any value higher than 150 yielded degraded performance, and the same applies to values less than 20 (AP maximum of 66\%). Lastly, we observed an increase in accuracy from 75.10\% to 77.18\% when we introduced varying spatial kernels to the convolutions after and before the transformer layers as mentioned in Section \ref{2d discussion}.

\noindent \textbf{Training Stride}
The stride controls how overlapped the data input is. The model's performance degraded noticeably when we remove the stride and take every unique window size. A jump of around 10\% in accuracy is usually observed on different models when an overlap of at least half to 75\% of the window size is introduced (e.g. overlap of 24 frames in a 32 window size input).

\noindent \textbf{Learning Rate \& Scheduling}
The model trains on multiple iterations per epoch, and it is possible to converge to a set of weights that produce acceptable results (71.23\% AP) within the first 3 epochs of training at a learning rate of $10^{-4}$. This learning rate is too large, however, starting with a learning rate of $10^{-5}$ proved to be too small. We used multiple scheduling techniques and starting/end points \cite{sgdr}. Starting with a learning rate of $10^{-4}$ and ending at $10^{-6}$ using both step scheduling and cosine annealing yielded the final models.

\vspace{-1mm}

\section{Conclusion} \label{conclusion}
\vspace{-4mm}
We introduce a novel transformers-based architecture for deep learning applications on radar frequency images, named RadarFormer. The main novelty lies in the reduction of total computing complexity and training/inference times of the original model by using transformers, a lightweight and efficient deep learning module. We also introduce a channel-chirp-time merging block that contributes to the reduction in computation complexity without compromising accuracy. Our multi-axis attention-based model produces state-of-the-art results while having significantly fewer parameters and inference time compared to the previous state-of-the-art method. The proposed models can pave the way for transformer-based research in radar frequency deep learning research.

 \bibliographystyle{splncs04}
 \bibliography{mybibliography}

\begin{thebibliography}{10}
\providecommand{\url}[1]{\texttt{#1}}
\providecommand{\urlprefix}{URL }
\providecommand{\doi}[1]{https://doi.org/#1}

\bibitem{autom-radar-class}
Angelov, A., Robertson, A., Murray-Smith, R., Fioranelli, F.: Practical
  classification of different moving targets using automotive radar and deep
  neural networks. IET Radar, Sonar \& Navigation  \textbf{12}(10),  1082--1089
  (2018). \doi{https://doi.org/10.1049/iet-rsn.2018.0103},
  \url{https://ietresearch.onlinelibrary.wiley.com/doi/abs/10.1049/iet-rsn.2018.0103}

\bibitem{radsegnet}
Bansal, K., Rungta, K., Bharadia, D.: Radsegnet: A reliable approach to radar
  camera fusion (2022). \doi{10.48550/ARXIV.2208.03849},
  \url{https://arxiv.org/abs/2208.03849}

\bibitem{kitti}
Behley, J., Garbade, M., Milioto, A., Quenzel, J., Behnke, S., Gall, J.,
  Stachniss, C.: {Towards 3D LiDAR-based semantic scene understanding of 3D
  point cloud sequences: The SemanticKITTI Dataset}. The International Journal
  on Robotics Research  \textbf{40}(8-9),  959--967 (2021).
  \doi{10.1177/02783649211006735}

\bibitem{radar-id}
Cao, P., Xia, W., Ye, M., Zhang, J., Zhou, J.: Radar-id: human identification
  based on radar micro-doppler signatures using deep convolutional neural
  networks. IET Radar, Sonar \& Navigation  \textbf{12}(7),  729--734 (2018).
  \doi{https://doi.org/10.1049/iet-rsn.2017.0511},
  \url{https://ietresearch.onlinelibrary.wiley.com/doi/abs/10.1049/iet-rsn.2017.0511}

\bibitem{vehicle-class}
Capobianco, S., Facheris, L., Cuccoli, F., Marinai, S.: Vehicle classification
  based on convolutional networks applied to fmcw radar signals. In: Leuzzi,
  F., Ferilli, S. (eds.) Traffic Mining Applied to Police Activities. pp.
  115--128. Springer International Publishing, Cham (2018)

\bibitem{hybrid1}
Dai, Z., Liu, H., Le, Q.V., Tan, M.: Coatnet: Marrying convolution and
  attention for all data sizes. In: Beygelzimer, A., Dauphin, Y., Liang, P.,
  Vaughan, J.W. (eds.) Advances in Neural Information Processing Systems
  (2021), \url{https://openreview.net/forum?id=dUk5Foj5CLf}

\bibitem{hybrid2}
D'Ascoli, S., Touvron, H., Leavitt, M.L., Morcos, A.S., Biroli, G., Sagun, L.:
  Convit: Improving vision transformers with soft convolutional inductive
  biases. In: Internation Conference on Machine Learning. pp. 2286--2296 (2021)

\bibitem{vit}
Dosovitskiy, A., Beyer, L., Kolesnikov, A., Weissenborn, D., Zhai, X.,
  Unterthiner, T., Dehghani, M., Minderer, M., Heigold, G., Gelly, S.,
  Uszkoreit, J., Houlsby, N.: An image is worth 16x16 words: Transformers for
  image recognition at scale (2020). \doi{10.48550/ARXIV.2010.11929},
  \url{https://arxiv.org/abs/2010.11929}

\bibitem{9000872}
Feng, D., Haase-Schütz, C., Rosenbaum, L., Hertlein, H., Gläser, C., Timm,
  F., Wiesbeck, W., Dietmayer, K.: Deep multi-modal object detection and
  semantic segmentation for autonomous driving: Datasets, methods, and
  challenges. IEEE Transactions on Intelligent Transportation Systems
  \textbf{22}(3),  1341--1360 (2021). \doi{10.1109/TITS.2020.2972974}

\bibitem{survey3}
Hassanin, M., Anwar, S., Radwan, I., Khan, F.S., Mian, A.: Visual attention
  methods in deep learning: An in-depth survey (2022).
  \doi{10.48550/ARXIV.2204.07756}, \url{https://arxiv.org/abs/2204.07756}

\bibitem{unetr}
Hatamizadeh, A., Tang, Y., Nath, V., Yang, D., Myronenko, A., Landman, B.,
  Roth, H.R., Xu, D.: Unetr: Transformers for 3d medical image segmentation.
  In: Proceedings of the IEEE/CVF Winter Conference on Applications of Computer
  Vision. pp. 574--584 (2022)

\bibitem{7780459}
He, K., Zhang, X., Ren, S., Sun, J.: Deep residual learning for image
  recognition. In: 2016 IEEE Conference on Computer Vision and Pattern
  Recognition (CVPR). pp. 770--778 (2016). \doi{10.1109/CVPR.2016.90}

\bibitem{mbconv}
Howard, A.G., Zhu, M., Chen, B., Kalenichenko, D., Wang, W., Weyand, T.,
  Andreetto, M., Adam, H.: Mobilenets: Efficient convolutional neural networks
  for mobile vision applications. CoRR  \textbf{abs/1704.04861} (2017),
  \url{http://arxiv.org/abs/1704.04861}

\bibitem{survey2}
Khan, S., Naseer, M., Hayat, M., Zamir, S.W., Khan, F.S., Shah, M.:
  Transformers in vision: A survey. ACM Comput. Surv.  \textbf{54}(10s) (sep
  2022). \doi{10.1145/3505244}, \url{https://doi.org/10.1145/3505244}

\bibitem{lahoud}
Lahoud, J., Cao, J., Khan, F.S., Cholakkal, H., Anwer, R.M., Khan, S., Yang,
  M.H.: 3d vision with transformers: A survey. arXiv preprint arXiv:2208.04309
  (2022)

\bibitem{hybrid3}
Li, Y., Wu, C.Y., Fan, H., Mangalam, K., Xiong, B., Malik, J., Feichtenhofer,
  C.: Mvitv2: Improved multiscale vision transformers for classification and
  detection. In: CVPR (2022)

\bibitem{radarfusion2019}
Lim, T.Y., Ansari, A., Major, B., Fontijne, D., Hamilton, M., Gowaikar, R.,
  Subramanian, S.: Radar and camera early fusion for vehicle detection in
  advanced driver assistance systems. {NeurIPS} Machine Learning for Autonomous
  Driving Workshop  (2019)

\bibitem{radical}
Lim, T.Y., Markowitz, S.A., Do, M.N.: Radical: A synchronized fmcw radar,
  depth, imu and rgb camera data dataset with low-level fmcw radar signals.
  IEEE Journal of Selected Topics in Signal Processing  \textbf{15}(4),
  941--953 (2021). \doi{10.1109/JSTSP.2021.3061270}

\bibitem{swin2}
Liu, Z., Hu, H., Lin, Y., Yao, Z., Xie, Z., Wei, Y., Ning, J., Cao, Y., Zhang,
  Z., Dong, L., Wei, F., Guo, B.: Swin transformer v2: Scaling up capacity and
  resolution. In: International Conference on Computer Vision and Pattern
  Recognition (CVPR) (2022)

\bibitem{swin1}
Liu, Z., Lin, Y., Cao, Y., Hu, H., Wei, Y., Zhang, Z., Lin, S., Guo, B.: Swin
  transformer: Hierarchical vision transformer using shifted windows. In:
  Proceedings of the IEEE/CVF International Conference on Computer Vision
  (ICCV) (2021)

\bibitem{sgdr}
Loshchilov, I., Hutter, F.: {SGDR}: Stochastic gradient descent with warm
  restarts. In: International Conference on Learning Representations (2017),
  \url{https://openreview.net/forum?id=Skq89Scxx}

\bibitem{stackedhourglass}
Newell, A., Yang, K., Deng, J.: Stacked hourglass networks for human pose
  estimation. In: Leibe, B., Matas, J., Sebe, N., Welling, M. (eds.) Computer
  Vision -- ECCV 2016. pp. 483--499. Springer International Publishing, Cham
  (2016)

\bibitem{carrada_model}
Ouaknine, A., Newson, A., P\'erez, P., Tupin, F., Rebut, J.: Multi-view radar
  semantic segmentation. In: Proceedings of the IEEE/CVF International
  Conference on Computer Vision (ICCV). pp. 15671--15680 (October 2021)

\bibitem{carrada}
Ouaknine, A., Newson, A., Rebut, J., Tupin, F., Pérez, P.: Carrada dataset:
  Camera and automotive radar with range- angle- doppler annotations. In: 2020
  25th International Conference on Pattern Recognition (ICPR). pp. 5068--5075
  (2021). \doi{10.1109/ICPR48806.2021.9413181}

\bibitem{hilo}
Pan, Z., Cai, J., Zhuang, B.: Fast vision transformers with hilo attention. In:
  NeurIPS (2022)

\bibitem{vtunet}
Peiris, H., Hayat, M., Chen, Z., Egan, G., Harandi, M.: A robust volumetric
  transformer for accurate 3d tumor segmentation. In: International Conference
  on Medical Image Computing and Computer-Assisted Intervention. pp. 162--172.
  Springer (2022)

\bibitem{yolo}
Redmon, J., Divvala, S.K., Girshick, R.B., Farhadi, A.: You only look once:
  Unified, real-time object detection. CoRR  \textbf{abs/1506.02640} (2015),
  \url{http://arxiv.org/abs/1506.02640}

\bibitem{radar_scenes_dataset}
Schumann, O., Hahn, M., Scheiner, N., Weishaupt, F., Tilly, J.F., Dickmann, J.,
  W{\"{o}}hler, C.: Radarscenes: {A} real-world radar point cloud data set for
  automotive applications. CoRR  \textbf{abs/2104.02493} (2021),
  \url{https://arxiv.org/abs/2104.02493}

\bibitem{UNETR++}
Shaker, A., Maaz, M., Rasheed, H., Khan, S., Yang, M.H., Khan, F.S.: Unetr++:
  Delving into efficient and accurate 3d medical image segmentation.
  arXiv:2212.04497  (2022)

\bibitem{7298594}
Szegedy, C., Liu, W., Jia, Y., Sermanet, P., Reed, S., Anguelov, D., Erhan, D.,
  Vanhoucke, V., Rabinovich, A.: Going deeper with convolutions. In: 2015 IEEE
  Conference on Computer Vision and Pattern Recognition (CVPR). pp.~1--9
  (2015). \doi{10.1109/CVPR.2015.7298594}

\bibitem{maxvit}
Tu, Z., Talebi, H., Zhang, H., Yang, F., Milanfar, P., Bovik, A., Li, Y.:
  Maxvit: Multi-axis vision transformer. ECCV  (2022)

\bibitem{10.3389/fnins.2022.851774}
Vogginger, B., Kreutz, F., López-Randulfe, J., Liu, C., Dietrich, R.,
  Gonzalez, H.A., Scholz, D., Reeb, N., Auge, D., Hille, J., Arsalan, M.,
  Mirus, F., Grassmann, C., Knoll, A., Mayr, C.: Automotive radar processing
  with spiking neural networks: Concepts and challenges. Frontiers in
  Neuroscience  \textbf{16} (2022). \doi{10.3389/fnins.2022.851774},
  \url{https://www.frontiersin.org/articles/10.3389/fnins.2022.851774}

\bibitem{monocular}
Wang, Y., Huang, Y.T., Hwang, J.N.: Monocular visual object 3d localization in
  road scenes. In: Proceedings of the 27th ACM International Conference on
  Multimedia. pp. 917--925. ACM (2019)

\bibitem{challenge}
Wang, Y., Hwang, J.N., Wang, G., Liu, H., Kim, K.J., Hsu, H.M., Cai, J., Zhang,
  H., Jiang, Z., Gu, R.: Rod2021 challenge: A summary for radar object
  detection challenge for autonomous driving applications. In: Proceedings of
  the 2021 International Conference on Multimedia Retrieval. pp. 553--559
  (2021)

\bibitem{rodnet}
Wang, Y., Jiang, Z., Gao, X., Hwang, J.N., Xing, G., Liu, H.: Rodnet: Radar
  object detection using cross-modal supervision. In: 2021 IEEE Winter
  Conference on Applications of Computer Vision (WACV). pp. 504--513 (2021).
  \doi{10.1109/WACV48630.2021.00055}

\bibitem{detr3d}
Wang, Y., Guizilini, V., Zhang, T., Wang, Y., Zhao, H., , Solomon, J.M.:
  Detr3d: 3d object detection from multi-view images via 3d-to-2d queries. In:
  The Conference on Robot Learning ({CoRL}) (2021)

\bibitem{earlyconv}
Xiao, T., Singh, M., Mintun, E., Darrell, T., Doll{\'a}r, P., Girshick, R.:
  Early convolutions help transformers see better. Advances in Neural
  Information Processing Systems  \textbf{34},  30392--30400 (2021)

\bibitem{sparse-single}
Yan, X., Gao, J., Li, J., Zhang, R., Li, Z., Huang, R., Cui, S.: Sparse single
  sweep lidar point cloud segmentation via learning contextual shape priors
  from scene completion. Proceedings of the AAAI Conference on Artificial
  Intelligence  \textbf{35}(4),  3101--3109 (May 2021).
  \doi{10.1609/aaai.v35i4.16419},
  \url{https://ojs.aaai.org/index.php/AAAI/article/view/16419}

\bibitem{lidarMN}
Ye, D., Chen, W., Zhou, Z., Xie, Y., Wang, Y., Wang, P., Foroosh, H.:
  Lidarmultinet: Unifying lidar semantic segmentation, 3d object detection, and
  panoptic segmentation in a single multi-task network (2022).
  \doi{10.48550/ARXIV.2206.11428}, \url{https://arxiv.org/abs/2206.11428}

\bibitem{hrformer}
Yuan, Y., Fu, R., Huang, L., Lin, W., Zhang, C., Chen, X., Wang, J.: Hrformer:
  High-resolution transformer for dense prediction. In: NeurIPS (2021)

\bibitem{9438625}
Yuan, Z., Song, X., Bai, L., Wang, Z., Ouyang, W.: Temporal-channel transformer
  for 3d lidar-based video object detection for autonomous driving. IEEE
  Transactions on Circuits and Systems for Video Technology  \textbf{32}(4),
  2068--2078 (2022). \doi{10.1109/TCSVT.2021.3082763}

\bibitem{raddet}
Zhang, A., Nowruzi, F.E., Laganiere, R.: Raddet: Range-azimuth-doppler based
  radar object detection for dynamic road users. In: 2021 18th Conference on
  Robots and Vision (CRV). pp. 95--102 (2021).
  \doi{10.1109/CRV52889.2021.00021}

\bibitem{panopticpolarnet}
Zhou, Z., Zhang, Y., Foroosh, H.: Panoptic-polarnet: Proposal-free lidar point
  cloud panoptic segmentation. In: Proceedings of the IEEE/CVF Conference on
  Computer Vision and Pattern Recognition (CVPR) (2021)

\end{thebibliography}
\end{document}